\documentclass[letterpaper]{article} % DO NOT CHANGE THIS
\usepackage{aaai23}  % DO NOT CHANGE THIS
\usepackage{times}  % DO NOT CHANGE THIS
\usepackage{helvet}  % DO NOT CHANGE THIS
\usepackage{courier}  % DO NOT CHANGE THIS
\usepackage[hyphens]{url}  % DO NOT CHANGE THIS
\usepackage{graphicx} % DO NOT CHANGE THIS
\usepackage{natbib}  % DO NOT CHANGE THIS AND DO NOT ADD ANY OPTIONS TO IT
\usepackage{caption} % DO NOT CHANGE THIS AND DO NOT ADD ANY OPTIONS TO IT
\usepackage{algorithm}
\usepackage{algorithmic}

\usepackage{microtype}
\usepackage{graphicx}
\usepackage{subfigure}
\usepackage{booktabs} % for professional tables
\usepackage{amsfonts}
\usepackage{multirow}
\usepackage{amsmath}
\usepackage{multirow}
\usepackage{lscape}
\usepackage{todonotes}
\usepackage{float} % appendix tables
\newcommand{\D}{\mathcal{D}}
\newcommand{\Dd}[1]{\mathcal{D}_{#1}}

\newtheorem{definition}{Definition}%[section]
\usepackage{newfloat}
\usepackage{listings}
\DeclareCaptionStyle{ruled}{labelfont=normalfont,labelsep=colon,strut=off} % DO NOT CHANGE THIS
\floatstyle{ruled}
\newfloat{listing}{tb}{lst}{}
\floatname{listing}{Listing}
\title{Monitoring Model Deterioration with Explainable Uncertainty Estimation via Non-parametric Bootstrap}
\author {
    Carlous Mougan\equalcontrib\textsuperscript{\rm 1},
    Dan Saattrup Nielsen\equalcontrib\textsuperscript{\rm 2}
}
\affiliations {
    \textsuperscript{\rm 1} University of Southampton, United Kingdom\\
    \textsuperscript{\rm 2} The Alexandra Institute, Denmark\\
    C.Mougan-Navarro@southampton.ac.uk, dan.nielsen@alexandra.dk
}
\begin{document}

\maketitle

\begin{abstract}
Monitoring machine learning models once they are deployed is challenging. It is even more challenging to decide when to retrain models in real-case scenarios when labeled data is beyond reach, and monitoring performance metrics becomes unfeasible. 
In this work, we use non-parametric bootstrapped uncertainty estimates and SHAP values to provide explainable uncertainty estimation as a technique that aims to monitor the deterioration of machine learning models in deployment environments, as well as determine the source of model deterioration when target labels are not available. Classical methods are purely aimed at detecting distribution shift, which can lead to false positives in the sense that the model has not deteriorated despite a shift in the data distribution.
To estimate model uncertainty we construct prediction intervals using a novel bootstrap method, which improves upon the work of \citet{kumar2012bootstrap}.
We show that both our model deterioration detection system as well as our uncertainty estimation method achieve better performance than the current state-of-the-art.
Finally, we use explainable AI techniques to gain an understanding of the drivers of model deterioration.
We release an open source Python package, \texttt{doubt}, which implements our proposed methods, as well as the code used to reproduce our experiments.
\end{abstract}
\section{Introduction}

Monitoring machine learning models in production is not an easy task. There are situations when the true label of the deployment data is available, and performance metrics can be monitored. But there are cases where it is not, and performance metrics are not so trivial to calculate once the model has been deployed. Model monitoring aims to ensure that a machine learning application in a production environment displays consistent behavior over time. 

Being able to explain or remain accountable for the performance or the deterioration of a deployed model is crucial, as a drop in model performance can affect the whole business process ~\citep{desiderataECB}, potentially having catastrophic consequences\footnote{The Zillow case is an example of consequences of model performance degradation in an unsupervised monitoring scenario, see
\url{https://edition.cnn.com/2021/11/09/tech/zillow-ibuying-home-zestimate/index.html} (Online accessed January 26, 2022).}. Once a deployed model has deteriorated, models are retrained using previous and new input data in order to maintain high performance. This process is called continual learning ~\cite{continual_learning} and it can be computationally expensive and put high demands on the software engineering system. Deciding when to retrain machine learning models is paramount in many situations.

Traditional machine learning systems assume that training data has been generated from a stationary source, but \textit{data is not static, it evolves}. This problem can be seen as a distribution shift, where the data distributions of the training set and the test set differ. Detecting distribution shifts has been a longstanding problem in the machine learning (ML) research community \cite{SHIMODAIRA2000227,sugiyama,sugiyama2,priorShift,learningSampleSelection,intuitionSampleSelection,varietiesSelectionBias,cortes2008sample,CorrectingSampleSelection,practicalFB}, as it is one of the main sources of model performance deterioration ~\citep{datasetShift}. Furthermore, data scientists in machine learning competitions claim that finding the train/validation split that better resembles the test (evaluation) distribution is paramount to winning a Kaggle competition ~\citep{howtowinKaggle}. 

However, despite the fact that a shift in data distribution can be a source of model deterioration, the two are not identical. Indeed, if we shift a random noise feature we have caused a change in the data distribution, but we should not expect the performance of a model to decline when evaluated on this shifted dataset. Thus, we emphasize here that our focus is on \textit{model deterioration} and not distribution shift, despite the correlation between the two.

Established ways of monitoring distribution shift when the real target distribution is not available are based on statistical changes either the input data ~\citep{continual_learning,DBLP:conf/nips/RabanserGL19} or on the model output \cite{garg2022leveraging}. These statistical tests correctly detect univariate changes in the distribution but are completely independent of the model performance and can therefore be too sensitive, indicating a change in the covariates but without any degradation in the model performance. This can result in false positives, leading to unnecessary model retraining. It is worth noting that several authors have stated the clear need to identify how non-stationary environments affect the behavior of models \citep{continual_learning}.

Aside from merely indicating that a model has deteriorated, it can in some circumstances be beneficial to identify the \textit{cause} of the model deterioration by detecting and explaining the lack of knowledge in the prediction of a model. Such explainability techniques can provide algorithmic transparency to stakeholders and to the ML engineering team ~\cite{desiderataECB,bhatt2021uncertainty,koh2020understanding,ribeiro2016why,sundararajan2017axiomatic}.

This paper's primary focus is on non-deep learning models and small to medium-sized tabular datasets, a size of data that is very common in the average industry, where, non-deep learning-based models achieve state-of-the-art results \cite{grinsztajn:hal-03723551,survey_DL_tabular,do_we_need_DL}.

Our contributions are the following:

\begin{enumerate}
    \item We develop a novel method that produces prediction intervals using bootstrapping with theoretical guarantees, which achieves better coverage than previous methods on eight real-life regression datasets from the UCI repository \cite{uci_data}.
    
    \item We use this non-parametric uncertainty estimation method to develop a machine learning monitoring system for regression models, which outperforms previous monitoring methods in terms of detecting deterioration of model performance.
    
    \item We use explainable AI techniques to identify the source of model deterioration for both entire distributions as a whole as well as for individual samples, where classical statistical indicators can only determine distribution differences. 

    \item We release an open source Python package, \texttt{doubt}, which implements our uncertainty estimation method and is compatible with all \texttt{scikit-learn} models \cite{pedregosa2011scikit}.
\end{enumerate}

%The rest of the paper is organized as follows: In Section~\ref{sec:relatedwork} we discuss related work and outline their differences to our contribution. Section~\ref{sec:methodology} introduces the construction of our prediction intervals and the choice of metrics we use to measure the performance of a model monitoring system. Section~\ref{sec:experiments} covers the experiments, results, and their corresponding discussion. Finally, in Section~\ref{sec:conclusion} we summarise the main conclusions of the paper.
\section{Related Work}\label{sec:relatedwork}

\subsection{Model Monitoring}

%From a software perspective, a machine learning model is another software component whose functionalities need to be tested before deployment. This approach has one structural weakness: \textit{data is not static, it evolves} ~\cite{continual_learning}. The conditions in which a system is developed can differ from deployment time for a range of reasons, from an evolutionary environment, a sampling bias at training time, or the inability to reproduce test conditions at training time ~\cite{datasetShift}.

Model monitoring techniques help to detect unwanted changes in the behavior of a machine learning application in a production environment. One of the biggest challenges in model monitoring is distribution shift, which is also one of the main sources of model degradation ~\cite{datasetShift,continual_learning}.

Diverse types of model monitoring scenarios require different supervision techniques. We can distinguish two main groups: Supervised learning and unsupervised learning. Supervised learning is the appealing one from a monitoring perspective, where performance metrics can easily be tracked. Whilst attractive, these techniques are often unfeasible as they rely either on having ground truth labeled data available or maintaining a hold-out set, which leaves the challenge of how to monitor ML models to the realm of unsupervised learning \cite{continual_learning}. Popular unsupervised methods that are used in this respect are the Population Stability Index (PSI) and the Kolmogorov-Smirnov test (K-S), all of which measure how much the distribution of the covariates in the new samples differs from the covariate distribution within the training samples. These methods are often limited to real-valued data, low dimensions, and require certain probabilistic assumptions~\citep{continual_learning,ShiftsData}.

%\citet{early_drift} presents an Early Drift Detection Method whose aim is to detect gradual drift, one of the most challenging types of drift \citep{continual_learning}, by calculating the distance between classification errors (number of examples between two classification errors). Our work differs from theirs in that we do not need the ground-truth labels of the test samples to monitor model deterioration.

Another approach suggested by \citet{shapTree} is to monitor the SHAP value contribution of input features over time together with decomposing the loss function across input features in order to identify possible bugs in the pipeline as well as distribution shift. This technique can account for previously unaccounted bugs in the machine learning production pipeline but fails to monitor the model degradation.

Is worth noting that prior work \cite{garg2022leveraging,jiang2021assessing} has focused on monitoring models either on out-of-distribution data or in-distribution data \cite{neyshabur2017exploring,neyshabur2018understanding}. Such a task, even if challenging, does not accurately represent the different types of data a model encounters in the wild. In a production environment, a model can encounter previously seen data (training data), unseen data with the same distribution (test data), and statistically new and unseen data (out-of-distribution data). That is why we focus our work on finding an unsupervised estimator that replicates the behavior of the model performance.

The idea of mixing uncertainty with dataset shift was introduced by \citet{trustUncertainty}. Our work differs from theirs, in that they evaluate uncertainty by shifting the distributions of their dataset, where we aim to detect model deterioration under dataset shift using uncertainty estimation. Their work is also focused on deep learning classification problems, while we estimate uncertainty using model agnostic regression techniques. Further, our contribution allows us to pinpoint the features/dimensions that are main causes of the model degradation.

\citet{garg2022leveraging} introduces a monitoring system for classification models, based on imposing thresholds on the softmax values of the model. Our method differs from theirs in that we work with regression models and not classification models, and that our method utilizes external uncertainty estimation methods, rather than relying on the model's own ``confidence'' (i.e., the outputted logits and associated softmax values).

\citet{DBLP:conf/nips/RabanserGL19}, presents a comprehensive empirical investigation of dataset shift, examining how dimensionality reduction and two-sample testing might be combined to produce a practical pipeline for detecting distribution shift in a real-life machine learning system. They show that the two-sample-testing-based approach performs best. This serves as a baseline comparison within our models, even if their idea is more focused on binary classification, whereas our works focus on building a regression indicator.

\subsection{Uncertainty}

Uncertainty estimation is being developed at a fast pace. Model averaging ~\cite{kumar2012bootstrap,gal2016dropout,lakshminarayanan2017simple,arnez2020} has emerged as the most common approach to uncertainty estimation. Ensemble and sampling-based uncertainty estimates have been successfully applied to many use cases such as detecting misclassifications \cite{unc_ood_class}, out-of-distribution inputs \citep{dangelo2021uncertaintybased}, adversarial attacks ~\cite{adversarialUncertainty,smith2018understanding}, automatic language assessments \cite{malinin2019uncertainty} and active learning \cite{kirsch2019batchbald}. In our work, we apply uncertainty to detect and explain model performance for seen data (train), unseen and identically distributed data (test), and statistically new and unseen data (out-of-distribution).

\citet{kumar2012bootstrap} introduced a non-parametric method to compute prediction intervals for any ML model using a bootstrap estimate, with theoretical guarantees. Our work is an extension of their work, where we take into account the model's variance in the construction of the prediction intervals. The result, as we will see in the experiments section, is that such intervals have better coverage in such high-variance scenarios.

\citet{barber2021predictive} recently introduced a new non-parametric method of creating prediction intervals, using the Jackknife+. Our method differs from theirs in that we are using general bootstrapped samples for our estimates, rather than leave-one-out estimates. In the experimental, we will see that the two methods perform similarly, but that our method is again more accurate in a high-variance scenario.

%Towards distribution shift benchmark detection there have been also recent developments. ~\cite{wilds} is a collection of datasets containing real-world distributional shifts for images and text, ~\cite{hendrycks1,hendrycks2,hendrycks3} proposed a set of datasets for benchmarking image classifier robustness to distribution shift, ~\cite{ShiftsData} proposed three datasets for evaluating uncertainty estimates and distribution shift on real industrial use cases, with the data in the three datasets belonging to different modalities (tabular, text and image).

%To the best of our knowledge, there are no data agnostic benchmark techniques to evaluate gradual model deterioration and most of the work relies on specific datasets.

%At a minimum, the outputs of the data monitoring subsystem are i) type of shift; ii) shift magnitude; iii) quantification of uncertainty. It is also desirable that explanations should be given for the shift occurring, such as

\section{Methodology}\label{sec:methodology}

\subsection{Evaluation of deterioration detection systems}\label{sec:evaluation-deterioration-systems}

The problem we are tackling in this paper is evaluating and accounting for model predictive performance deterioration. To do this, we simulate a distribution shift scenario in which we \textit{have} access to the true labels, which we can use to measure the model deterioration and thus evaluate the monitoring system. A naive simulation in which we simply manually shift a chosen feature of a dataset would not be representative, as the associated true labels could have changed if such a shift happened "in the wild".

Therefore, we propose the following alternative approach. Starting from a real-life dataset $\D$ and a numerical feature $F$ of $\D$, we sort the data samples of $\D$ by the value of $F$, and split the sorted $\D$ in three equally sized sections: $\{\Dd{below},\Dd{tr},\Dd{upper}\} \subseteq\D$. The model is then fitted to the middle section ($\Dd{tr}$) and evaluated on all of $\D$. The goal of the monitoring system is to input the model, the labeled data segment $\Dd{tr}$ and a sample of unlabelled data $\mathcal{S}\subseteq\D$, and output a ``monitoring value'' which behaves like the model's performance on $\mathcal{S}$. Such a prediction will thus have to take into account the training performance, generalization performance as well as the out-of-distribution performance of the model.

In the experimental section, we compare our monitoring technique to several other such systems. To enable comparison between the different monitoring systems, we standardize all monitoring values as well as the performance metrics of the model. From these standardized values, we can now directly measure the goodness-of-fit of the model monitoring system by computing the absolute difference between its (standardized) monitoring values and the (standardized) ground truth model performance metrics. Our chosen evaluation method is very similar to the one used by \citet{garg2022leveraging}. They focus on classification models and their systems output estimates of the model's accuracy on the dataset. They evaluate these systems by computing the absolute difference between the system's accuracy estimate and the actual accuracy that the model achieves on the dataset.

As we are working with regression models in this paper, we will only operate with a single model performance metric: mean squared error. We will introduce our monitoring system, which is based on an uncertainty measure, and will compare our monitoring system against statistical tests based on input data or prediction data. In that section, we will also compare our uncertainty estimation method to current state-of-art uncertainty estimation methods.

\subsection{Uncertainty estimation}\label{sec:uncertaintyestimation}

In order to estimate uncertainty in a general way for all machine learning models, we use a non-parametric regression technique, which is an improvement of the technique introduced by \cite{kumar2012bootstrap}. This method aims at determining prediction intervals for outputs of general non-parametric regression models using bootstrap methods.

Setting $d\in\mathbb{N}$ to be the dimension of the feature space, we assume that the true model $y\colon\mathbb{R}^d\to\mathbb{R}$ is of the form $y(x) = \delta(x) + \varepsilon(x)$, where $\delta\colon\mathbb{R}^d\to\mathbb{R}$ is a deterministic and continuously differentiable function, and the \textbf{observation noise} $\varepsilon\colon\mathbb{R}^d\to\mathbb{R}$ is a uniform random field such that $\varepsilon(x_1),\dots,\varepsilon(x_t)$ are iid for any $x_1,\dots,x_t\in\mathbb{R}^d$, have zero mean and finite variance. We will assume that we have a data sample $X$ of size $N$, as well as a \textit{convergent estimator} $\hat{\delta}^{(n)}$ of $\delta$, meaning the following:

\begin{definition}
    Let $\hat{\delta}^{(n)}\colon\mathbb{R}^d\to\mathbb{R}$ be a function for every $n\in\mathbb{N}$. We then say that $\hat{\delta}^{(n)}$ is a \textbf{convergent estimator} of a function $\delta\colon\mathbb{R}^d\to\mathbb{R}$ if:
    \begin{enumerate}
        \item\label{def:convergentestimator_continuity} $\hat{\delta}^{(n)}$ is deterministic and continuous, for all $n\in\mathbb{N}$.
        \item\label{def:convergentestimator_convergence} There is a function $\hat{\delta}\colon\mathbb{R}^d\to\mathbb{R}$ such that $\hat{\delta}^{(n)}$ converges pointwise to $\hat{\delta}$ as $n\to\infty$.
        %\item There is a \textbf{bias function} $\beta\colon\mathbb{R}^d\to\mathbb{R}$ such that $\mathbb{E}_x[(\hat{\delta}^{(n)}(x) - \delta(x))^2]\to\beta(x)$ as $n\to\infty$.
    \end{enumerate}
\end{definition}

We define an associated \textbf{bias function} $\beta(x):=\delta(x)-\hat{\delta}(x)$. Note that in \citet{kumar2012bootstrap} they assumed that $\mathbb{E}[(\hat{\delta}^{(n)}(x)-\delta(x))^2]\to 0$ for $n\to\infty$, effectively meaning that the candidate model would be able to perfectly model the underlying distribution given enough data. It turns out that their method does not require this assumption, as we will see below. Aside from removing this assumption, the primary difference between our approach and \citet{kumar2012bootstrap} is that our approach extends the latter by maintaining good coverage in a high-variance situation, as we will also see below. We start by rewriting the equation for the true model as follows:
\begin{align}
    \label{eq:modelvariance}
    y(x) &= \delta(x) + \varepsilon(x) \\
    &= \hat{\delta}^{(N)}(x) + \beta(x) + \varepsilon_v^{(N)}(x) + \varepsilon(x),
\end{align}

where $\varepsilon_v^{(N)}(x) := \delta(x) - \beta(x) - \hat{\delta}^{(N)}(x)$ is the \textbf{model variance noise}. Note that
\begin{equation}
    \varepsilon_v^{(n)}(x) = \hat{\delta}(x) - \hat{\delta}^{(n)}(x)\to 0 \quad\textrm{as}\quad n\to\infty.
\end{equation}

To produce correct prediction intervals we thus need to estimate the distribution of the observation noise, bias and model variance noise.
 
\subsubsection{Estimating Model Variance Noise}\label{sec:modelvariancenoise}
To estimate the model variance noise term $\varepsilon_v^{(N)}(x)$ we adapt the technique in \citet{kumar2012bootstrap} to our scenario, using a bootstrap estimate. Concretely, we bootstrap our dataset $B>0$ times, fitting our model on each of the bootstrapped samples $X_b$ and generating bootstrapped estimates $\overline{\delta}_b^{(N)}(x)$ for every $b<B$. 
Centering the bootstrapped predictions as $m_b^{(N)}(x) := \mathbb{E}_b[\bar{\delta}_b^{(N)}]-\bar{\delta}_b^{(N)}$, we have that

\begin{align}
    \label{eq:modelvarianceestimate}
    \mathbb{D}_b[m_b^{(N)}(x)] &= \mathbb{D}_b[\mathbb{E}_b[\bar{\delta}_b^{(N)}(x)] - \bar{\delta}_b^{(N)}(x)] \\
    &\to \mathbb{D}_X[\hat{\delta}(x)-\bar{\delta}^{(N)}(x)] \\
    &= \mathbb{D}_X[\varepsilon_v^{(N)}(x)]
\end{align}

as $B\to\infty$, giving us our estimate of the model variance noise.

\subsubsection{Estimating Bias and Observation Noise}\label{sec:biasobsnoise}
We next have to estimate the bias $\beta(x)$ and the observation noise $\varepsilon(x)$. By rewriting (\ref{eq:modelvariance}) we get that that
\begin{equation}
    \beta(x)+\varepsilon(x)=y(x)-\hat{\delta}^{(N)}(x)-\varepsilon_v^{(N)}(x),
\end{equation}

so since we already have an estimate for $\varepsilon_v^{(N)}(x)$, it remains to estimate the residual $y(x)-\hat{\delta}^{(N)}(x)$. 
In \citet{kumar2012bootstrap} this was estimated purely using the training residuals without using any bootstrapping, whereas our approach will estimate the expected value of this residual via a bootstrap estimate, by using bootstrapped validation residuals $y(x)-\bar{\delta}_b^{(N)}(x)$, where $x$ is \textit{not} in the $b$'th bootstrap sample $X_b$. Concretely, we have that
\begin{align}
    &\mathbb{D}_{(b, x\in X\backslash X_b)}[y(x)-\bar{\delta}_b^{(N)}(x)]\\
    &\to \mathbb{D}_{(X, x\notin X)}[y(x)-\hat{\delta}^{(N)}(x)]
\end{align}

as $B\to\infty$. An initial estimate is thus
\begin{align}
    &\mathbb{D}_{(X, x\notin X)}[\beta(x)+\varepsilon(x)]] \\
    \label{eq:initialbiasobsnoise}
    &\approx \mathbb{D}_{(b, x\in X\backslash X_b)}[y(x)-\bar{\delta}_b^{(N)}(x) - m_b^{(N)}(x)]
\end{align}

Denote (\ref{eq:initialbiasobsnoise}) by $\texttt{valError}_b^{(N)}$. The problem with this approach is that the resulting prediction intervals arising from these validation errors are going to be too wide, as the bootstrap samples only contain on average 63.2\% of the samples in the original dataset \citep{statisticallearning}, causing the model to have artificially large validation residuals. To fix this, we follow the approach in \citet{statisticallearning} and use the 0.632+ bootstrap estimate instead, defined as follows. We start by defining the \textbf{no-information error rate}
\begin{equation}
    \texttt{noInfoError}^{(N)} := \frac{1}{n^2}\sum_{i=1}^n\sum_{j=1}^n(y(x_i)-\hat{\delta}(x_j))^2,
\end{equation}

corresponding to the mean-squared error if the inputs and outputs were independent. Next, define the associated training residuals $\texttt{trainError}_b^{(N)}$ as:
\begin{equation}
    \mathbb{D}_{(b, x\in X_b)}[y(x)-\bar{\delta}_b^{(N)}(x) - (\mathbb{E}_b[\bar{\delta}_b^{(N)}(x)] - \bar{\delta}_b^{(N)}(x))]].
\end{equation}

Combining these two, we set the \textbf{relative overfitting rate} $\texttt{overfittingRate}_b^{(N)}$ to be:
\begin{equation}
    \frac{\texttt{valError}_b^{(N)}-\texttt{trainError}_b^{(N)}}{\texttt{noInfoError}-\texttt{trainError}_b^{(N)}}.
\end{equation}

This gives us a convenient number between $0$ and $1$, denoting how much our model is overfitting the dataset. From this, we define the \textbf{validation weight} $\texttt{valWeight}_b^{(N)}$ as:
\begin{equation}
    \frac{0.632}{1-(1-0.632)\times\texttt{overfittingRate}_b^{(N)}},
\end{equation}

which denotes how much we should weigh the validation error over the training error. In case of no overfitting, we get that $\texttt{valWeight}_b^{(N)}=0.632$ and this reduces to the standard $0.632$ bootstrap estimate \cite{statisticallearning}, whereas in case of severe overfitting the weight becomes $1$ and thus only prioritizes the validation error.

Our final estimate of $\beta(x)+\varepsilon(x)$ is thus
\begin{align}
    \label{eq:biasobsnoise}
    &\mathbb{D}_X[\beta(x)+\varepsilon(x)] \approx \mathbb{D}_b[o_b^{(N)}],
\end{align}

where
\begin{align*}
    o_b^{(N)} :=\ &(1-\texttt{valWeight}_b^{(N)})\times\texttt{trainError}_b^{(N)} + \\ &\texttt{valWeight}_b^{(N)}\times\texttt{valError}_b^{(N)}.
\end{align*}

Note that this estimate is only an aggregate and is not specific to any specific value of $x$, as opposed to the model variance estimate in equation (\ref{eq:modelvarianceestimate}).

\subsubsection{Prediction Interval Construction}
Calculating the estimate of the prediction interval is then a matter of joining the results from the section of model variance noise and bias observation noise, in the same way as in \citet{kumar2012bootstrap}. As the estimate of $\beta(x)+\varepsilon(x)$ does not depend on any new sample, we can pre-compute this in advance by bootstrapping $B$ samples $X_b$, fit our model to each and calculate the $o_b^{(N)}$ using Equation (\ref{eq:biasobsnoise}). Now, given a new data point $x_0$ and $\alpha\in(0,1)$, we can estimate an $\alpha$ prediction interval around $x_0$ as follows. We again bootstrap $B$ samples $X_b$, fit our model to each and calculate the $m_b^{(N)}(x_0)$ values. Next, we form the set $C^{(N)}(x_0) := \{m_b^{(N)}(x_0) +o_b^{(N)} \mid b<B\}$, and our interval is then $(\texttt{start}, \texttt{end})$, where
\begin{align}
    &\texttt{start} := \hat{\delta}^{(N)}(x_0) - q_{\tfrac{\alpha}{2}}(C^{(N)}(x_0)) \\ 
    &\texttt{end} := \hat{\delta}^{(N)}(x_0) + q_{1-\tfrac{\alpha}{2}}(C^{(N)}(x_0)),
\end{align}

with $q_\xi(C^{(N)}(x_0))$ being the $\xi$'th quantile of $C^{(N)}(x_0)$.

\subsection{Detecting the source of uncertainty/model deterioration}
Using uncertainty as a method to monitor the performance of an ML model does not provide any information on \textit{what} features are the cause of the model degradation, only a goodness-of-fit to the model performance. We propose to solve this issue with the use of Shapley values. 

We start by fitting a model $f_{\theta}$ to the training data, $X^{\text{train}}$. We next shift the test data by five standard deviations (call the shifted data $X^{\text{ood}}$) and compute uncertainty estimates $Z$ of $f_{\theta}$ on $X^{\text{ood}}$. We next fit a second model $g_{\psi}$ on $(X^{\text{ood}})$ to predict the uncertainty estimate $Z$, and compute the associated Shapley values \cite{shapTree} of $g_{\psi}$. These Shapley values thus signify which features are the ones contributing the most to the uncertainty values. With the correlation between uncertainty values and model deterioration that we hope to conclude from the experiment described in the experimental section, this thus also provides us with a plausible cause of the model deterioration, if deterioration has taken place. Particularly, this methodology can be extended to large-scale datasets and deep learning-based models.
\section{Experiments}
\label{sec:experiments}
Our experiments have been
organized into three main groups: Firstly, we compare our non-parametric bootstrapped estimation method with the previous state-of-the-art, \citet{kumar2012bootstrap} and \citet{barber2021predictive}. Secondly, we assess the performance of our proposed uncertainty method for monitoring the performance of a machine learning model. And then, we evaluate the usability of the explainable uncertainty for identifying the features that are driving model degradation in local and global scenarios. In the main body of the paper, we present the results over several real-world datasets in the appendix we provide the experiments on synthetic datasets that exhibits, non-linear and linear behavior.

\subsection{Uncertainty method comparison}
To demonstrate the accuracy of our prediction intervals introduced in the uncertainty estimation section, we compare the coverage of the intervals with the NASA method from \citet{kumar2012bootstrap} on eight regression datasets from the UCI repository \citep{uci_data}. The statistics of these datasets can be seen in Table~\ref{tab:uncertaintydatasets}.%\footnote{All the datasets can be found at \url{https://archive.ics.uci.edu/ml/datasets} with \texttt{<name>} being the dataset name in Table~\ref{tab:uncertaintydatasets}, including spaces.}

\begin{table}[ht]
\begin{center}
\begin{tabular}{l|c|c}
    Dataset & \# Samples & \# Features\\
    \hline
    Airfoil Self-Noise             & 1,503            & 5  \\
    Bike Sharing                   & 17,379           & 16 \\
    Concrete Compressive Strength  & 1,030            & 8  \\
    QSAR Fish Toxicity             & 908              & 6  \\
    Forest Fires                   & 517              & 12 \\
    Parkinsons                     & 5,875            & 22 \\
    Power Plant                    & 9,568            & 4  \\
    Protein                        & 45,730           & 9 \\

\end{tabular}
\caption{Statistics of the regression datasets used in this paper.}\label{tab:uncertaintydatasets}
\end{center}
\end{table}

We split each of the eight datasets into a 90/10 train/test split, uniformly at random. Next, we fit a linear regression, a decision tree, and a gradient boosting decision tree on the training split. We chose these three models to have an example of a model with large bias (the linear regression model), a model with large variance (the decision tree model), and an intermediate model that achieves state-of-the-art performance in many tasks, the gradient boosting model. We will use the \texttt{xgboost} \cite{xgboost} implementation of the gradient boosting model. After fitting the three models we compute $\alpha$-prediction intervals for $\alpha\in\{0.75, 0.76, \dots, 0.99\}$, using our ``Doubt'' prediction intervals, the ``NASA'' prediction intervals from \cite{kumar2012bootstrap} as well as the ``MAPIE'' prediction intervals from \cite{barber2021predictive}, the latter implemented with the \texttt{MAPIE} package\footnote{\url{https://github.com/scikit-learn-contrib/MAPIE}}. We can then compare the coverage of the three methods on the eight test sets.

As the goal of an $\alpha$ prediction interval is to have a coverage of $\alpha$, we can measure the performance of a prediction interval system by reporting the absolute difference between the actual coverage of the interval and this ideal coverage $\alpha$. In Table~\ref{tab:uncertaintyresults} we report the mean and standard deviations of these absolute differences, for each of the three model architectures. We have performed pairwise two-tailed paired t-tests on all absolute differences, and the best-performing prediction interval methods are marked in bold for each model architecture.

We see (cf. Table~\ref{tab:uncertaintyresults}) that there is no significant difference between the three methods in the high bias case with the linear regression model. In the case of the XGBoost model, a model with higher variance, both the MAPIE and Doubt methods outperform the NASA method, but there is no significant difference between the MAPIE method and the Doubt method in this case. In the high-variance scenario with the decision tree, however, the Doubt intervals achieve significantly better coverage than both of the other two methods.

\begin{table*}[ht]
\begin{center}
\begin{tabular}{l|c|c|c}
    Model & Linear Regression & XGBoost & Decision Tree \\
    \hline
    NASA & $\mathbf{3.854 \pm 5.380}$ & $20.216 \pm 12.405$ & $20.669 \pm 9.771$\\
    MAPIE & $\mathbf{4.207 \pm 4.755}$ & $\mathbf{5.264 \pm 4.258}$ & $7.788 \pm 4.782$\\
    Doubt & $\mathbf{3.917 \pm 4.870}$ & $\mathbf{4.861 \pm 3.938}$ & $\mathbf{5.137 \pm 3.984}$
\end{tabular}
\caption{A comparison of different prediction interval methods, where the metric used is the mean absolute deviation from the ideal coverage (lower is better), with its associated standard deviation. Here NASA is the method described in  \citet{kumar2012bootstrap}, MAPIE is the method from \citet{barber2021predictive} and Doubt is our method. The best results for each model architecture are shown in bold.}\label{tab:uncertaintyresults}
\end{center}
\end{table*}

\subsection{Evaluating model deterioration}

The scenario we are addressing is characterized by regression data sets that have statistically seen data (train data), iid statistically unseen data (test data), and out-of-distribution data. Following the open data for reproducible research guidelines described in \citet{the_turing_way_community_2019_3233986} and for measuring the performance of the proposed methods, we have used eight open-source datasets (cf. Table~\ref{tab:uncertaintydatasets}) for an empirical comparison coming from the UCI repository \citep{uci_data}. As described in the methodology, in order to benchmark our algorithm we, for each feature $F$ in each dataset $\D$, sort $\D$ according to $F$ and split $\D$ into three equally sized sections $\{\Dd{below},\Dd{tr},\Dd{upper}\} \subseteq\D$. We then train the model on $\Dd{tr}$ and test the performance of all of $\D$. In this way we obtain a mixture of train, test, and out-of-distribution data, allowing us to evaluate our monitoring techniques in all three scenarios.

\begin{figure}[ht]
  \centering
  \includegraphics[width=1.1\columnwidth]{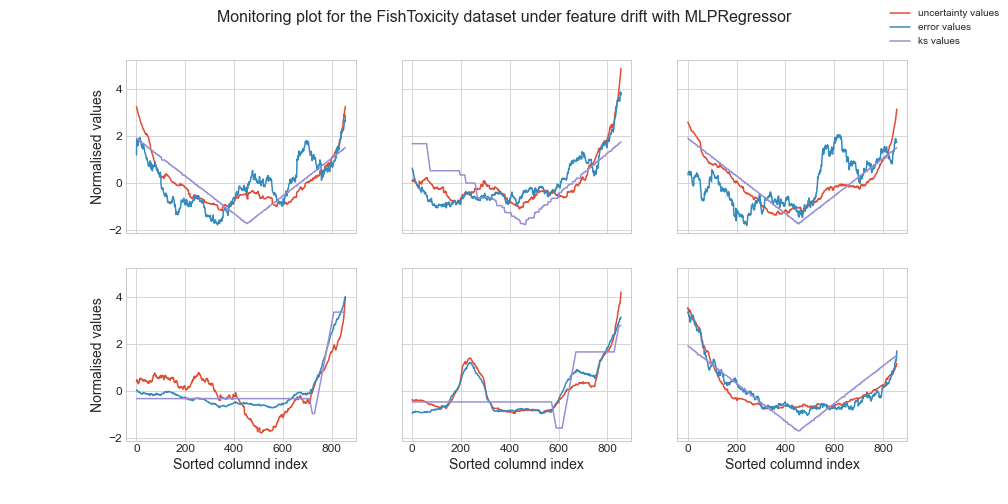}
  \caption{Comparison of different model degradation detection methods for the Fish Toxicity dataset. Each of the plots represents an independent experiment where each of the six features has been shifted, using the method described in the methodology section. Doubt achieves a better goodness-of-fit than previous statistical methods. A larger version of this figure can be found in Appendix (cf. Figure \ref{fig:distributionBIG}).}\label{fig:distribution}
\end{figure}

In evaluating a monitoring system we need to make a concrete choice of the sampling method to get the unlabelled data $\mathcal{S}\subseteq\D$. We are here using a rolling window of fifty samples, which has the added benefit of giving insight into the performance of the monitoring system on each of the three sections $\Dd{lower}$, $\Dd{tr}$ and $\Dd{upper}$ (cf. Figure~\ref{fig:distribution}).

We compare our monitoring system using the uncertainty estimation method against: $(i)$ two classical statistical methods on input data: the Kolmogorov-Smirnov test statistic (K-S) and the Population-Stability Index (PSI) \citep{continual_learning}, $(ii)$ a Kolmogorov-Smirnov statistical test on the predictions between train and test \cite{garg2022leveraging} that we denominate prediction shift and $(iii)$ the previous state-of-the-art  uncertainty estimate MAPIE. We evaluate the monitoring systems on a variety of model architectures: generalized linear models, tree-based models as well as neural networks.

The average performance across all datasets can be found in Table~\ref{tab:scores}.\footnote{See the appendix for a more detailed table.} From these we can see that our methods outperform K-S and PSI in all cases except for the Random Forest case, where our method is still on par with the best method, in that case, K-S. We have included a table with each dataset and all the estimators (cf. Table~\ref{tab:appendix} in the appendix), where it can be seen that both K-S and PSI easily identify a shift in the distribution but fail to detect when the model performance degrades, giving too many false positives.

\begin{table*}[ht]
\begin{center}
\begin{tabular}{l|cccccc}
    Method & Linear Regression & Poisson & Decision Tree & Random Forest & Gradient Boosting & MLP \\
    \hline
    PSI          & $0.87 \pm 0.08$          & $0.93 \pm 0.08$          & $0.97 \pm 0.10$          & $0.95 \pm 0.08$          & $0.95 \pm 0.08$          & $0.84 \pm 0.16$ \\
    K-S          & $0.81 \pm 0.10$          & $0.94 \pm 0.20$          & $0.52 \pm 0.12$          & $\mathbf{0.50 \pm 0.12}$ & $0.61 \pm 0.19$          & $0.72 \pm 0.22$ \\
    Prediction Shift & $0.86 \pm 0.13$          & $1.00 \pm 0.15$          & $0.80 \pm 0.14$          & $0.73 \pm 0.18$          & $0.75 \pm 0.20$          & $0.74 \pm 0.22$ \\
    MAPIE        & $0.77 \pm 0.10$          & $0.83 \pm 0.18$          & $0.60 \pm 0.16$          & $0.86 \pm 0.15$          & $0.73 \pm 0.18$          & $0.74 \pm 0.38$ \\
    Doubt        & $\mathbf{0.71 \pm 0.14}$ & $\mathbf{0.79 \pm 0.14}$ & $\mathbf{0.49 \pm 0.10}$ & $0.74 \pm 0.18$          & $\mathbf{0.58 \pm 0.23}$ & $\mathbf{0.68 \pm 0.38}$ \\
\end{tabular}
\caption{Performance of model monitoring systems for model deterioration for a variety of model architectures on eight regression datasets from the UCI repository \cite{uci_data}. The scores are the means and standard deviations of the absolute deviation from the true labels on $\Dd{lower}$ and $\Dd{upper}$ (lower is better). K-S and PSI are the monitoring systems obtained by computing the Kolmogorov-Smirnov test values and the Population Stability Index, respectively, Prediction Shift is the statistical comparison of the model prediction,  and Doubt is our method. The best results for each model architecture are shown in bold. See Table \ref{tab:appendix} in the Appendix for all the raw scores.}
\label{tab:scores}
\end{center}
\end{table*}

\subsection{Detecting the source of uncertainty}\label{sec:unceratainty_source}

For this experiment, we make use of two datasets: the synthetic one (on the Appendix) and the popular and intuitive House Prices regression dataset\footnote{\url{https://www.kaggle.com/c/house-prices-advanced-regression-techniques}}, whose goal is to predict the selling price of a given property. We select two of the features that are the most correlated with the target, \texttt{GrLivArea} and \texttt{TotalBsmtSF}, being the size of the living and base areas in square feet, respectively. We also create a new feature of random noise, to have an example of a feature with minimum correlation with the target. A model deterioration system should therefore highlight the \texttt{GrLivArea} and \texttt{TotalBsmtSF} features, and \textit{not} highlight the random features.

Concretely, we compute an estimation of the Shapley values using TreeSHAP, which is an efficient estimation approach values for tree-based models \cite{shapTree}, that allows for this second model to identify the features that are the source of the uncertainty, and thus also provide an indicator for what features might be causing the model deterioration.

We fitted an MLP on the training dataset, which achieved a $R^2$ value of $0.79$ on the validation set. We then shifted all three features by five standard deviations and trained a gradient boosting model on the uncertainty values of the MLP on the validation set, which achieves a good fit (an $R^2$ value of $0.94$ on the hold-out set of the validation). We then compare the SHAP values of the gradient boosting model with the PSI and K-S statistics for the individual features.

%In this experiment we shift the first \textbf{three} features of the Airfoil dataset\footnote{\url{https://archive.ics.uci.edu/ml/datasets/Airfoil+Self-Noise}} by five standard deviations,

\begin{figure}[ht]
  \centering
  \includegraphics[width=1.1\linewidth]{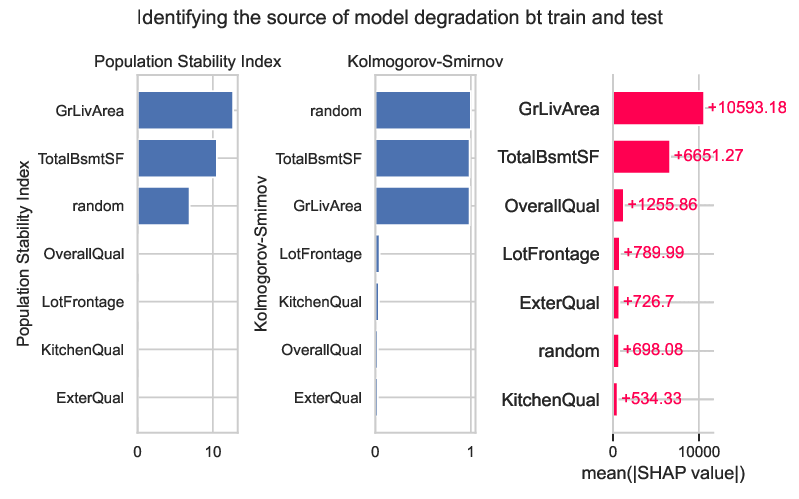}
  \caption{Global comparison of different distribution shift detection methods. Statistical methods correctly indicate that there exists a distribution shift in the shifted data. Shapley values indicate the contribution of each feature to the drop in predictive performance of the model.}\label{fig:shap}
\end{figure}

In Figure~\ref{fig:shap}, classical statistics and SHAP values to detect the source of the model deterioration are compared. We see that the PSI and K-S value correctly capture the shift in each of the three features (including the random noise). On the other hand, our SHAP method highlights the two substantial features (\texttt{GrLivArea} and \texttt{TotalBsmtSF}) and correctly does not assign a large value to the random feature, despite the distribution shift.

\begin{figure}[ht]
  \centering
  \includegraphics[width=1.1\linewidth]{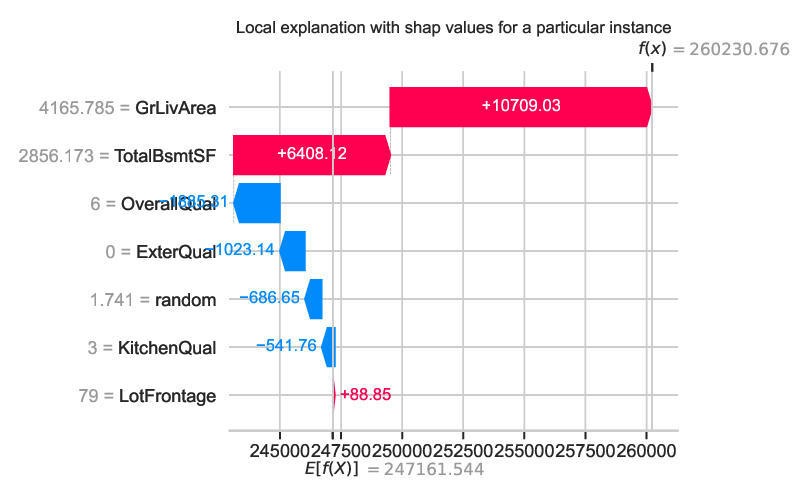}
  \caption{Individual explanation that displays the source of uncertainty for one instance. The previous method allowed only for comparison between distributions, now with explainable uncertainty, we are able to account for individual instances. In red, features pushing the uncertainty prediction higher are shown; in blue, those pushing the uncertainty prediction lower.}\label{fig:local_shap}
\end{figure}

Figure~\ref{fig:local_shap} shows features contributing to pushing the model output from the base value to the model output. Features pushing the uncertainty prediction higher are shown in red, and those pushing the uncertainty prediction lower are in blue \citep{lundberg2018,lundberg2020local2global,lundberg2017unified}. From these values, we can, at a local level, also identify the two features (\texttt{GrLivArea} and \texttt{TotalBsmtSF}) causing the model deterioration in this case.

\section{Conclusion}
\label{sec:conclusion}

In this work, we have provided methods and experiments to monitor and identify machine learning model deterioration via non-parametric bootstrapped uncertainty estimation methods, and use explainability techniques to explain the source of the model deterioration.

Our monitoring system is based on a novel uncertainty estimation method, which produces prediction intervals with theoretical guarantees and which achieves better coverage than the current state-of-the-art. The resulting monitoring system was shown to more accurately detect model deterioration than methods using classical statistics. Finally, we used SHAP values in conjunction with these uncertainty estimates to identify the features that are driving the model deterioration at both a global and local level and qualitatively showed that these more accurately detect the source of the model deterioration compared to classical statistical methods.

\textbf{Limitations:}We emphasize here that due to computationally limitations, we have only benchmarked on datasets of relatively small to medium size (cf. Table~\ref{tab:uncertaintydatasets}), and further work needs to be done to see if these results are also valid for datasets of significantly larger size. This work also focused on tabular data and non-deep learning models.

\subsection*{Reproducibility Statement}\label{sec:reproducibility}
To ensure reproducibility of our results, we make the data, data preparation routines, code repositories, and methods publicly available\footnote{\url{https://github.com/cmougan/MonitoringUncertainty}}.
Our novel uncertainty methods are included in the open-source Python package \texttt{doubt}\footnote{\url{https://github.com/saattrupdan/doubt}.}.
%Note that we do not perform any hyperparameter tuning throughout our work. Instead, we use default \texttt{scikit-learn} parameters.
We describe the system requirements and software dependencies of our experiments. Our experiments were run on an 8 vCPU server with 60 GB RAM.

\subsection*{Acknowledgements}\label{sec:acknowledgements}
This work has received funding by the European Union's Horizon 2020 research and
innovation programme under the Marie Sk\l odowska-Curie Actions (grant agreement number
860630) for the project : "NoBIAS - Artificial Intelligence without Bias". Furthermore,
this work reflects only the authors' view and the European Research Executive Agency
(REA) is not responsible for any use that may be made of the information it contains.

\bibliography{main}

\begin{thebibliography}{49}
\providecommand{\natexlab}[1]{#1}

\bibitem[{Arnez et~al.(2020)Arnez, Espinoza, Radermacher, and
  Terrier}]{arnez2020}
Arnez, F.; Espinoza, H.; Radermacher, A.; and Terrier, F. 2020.
\newblock A Comparison of Uncertainty Estimation Approaches in Deep Learning
  Components for Autonomous Vehicle Applications.
\newblock In Espinoza, H.; McDermid, J.~A.; Huang, X.; Castillo{-}Effen, M.;
  Chen, X.~C.; Hern{\'{a}}ndez{-}Orallo, J.; h{\'{E}}igeartaigh, S.~{\'{O}}.;
  and Mallah, R., eds., \emph{Proceedings of the Workshop on Artificial
  Intelligence Safety 2020 co-located with the 29th International Joint
  Conference on Artificial Intelligence and the 17th Pacific Rim International
  Conference on Artificial Intelligence {(IJCAI-PRICAI} 2020), Yokohama, Japan,
  January, 2021}, volume 2640 of \emph{{CEUR} Workshop Proceedings}.
  CEUR-WS.org.

\bibitem[{Arnold et~al.(2019)Arnold, Bowler, Gibson, Herterich, Higman,
  Krystalli, Morley, O'Reilly, Whitaker
  et~al.}]{the_turing_way_community_2019_3233986}
Arnold, B.; Bowler, L.; Gibson, S.; Herterich, P.; Higman, R.; Krystalli, A.;
  Morley, A.; O'Reilly, M.; Whitaker, K.; et~al. 2019.
\newblock The turing Way: a handbook for reproducible data science.
\newblock \emph{Zenodo}.

\bibitem[{Barber et~al.(2021)Barber, Candes, Ramdas, and
  Tibshirani}]{barber2021predictive}
Barber, R.~F.; Candes, E.~J.; Ramdas, A.; and Tibshirani, R.~J. 2021.
\newblock Predictive inference with the jackknife+.
\newblock \emph{The Annals of Statistics}, 49(1): 486--507.

\bibitem[{Bhatt et~al.(2021)Bhatt, Antor{\'a}n, Zhang, Liao, Sattigeri,
  Fogliato, Melan{\c{c}}on, Krishnan, Stanley, Tickoo
  et~al.}]{bhatt2021uncertainty}
Bhatt, U.; Antor{\'a}n, J.; Zhang, Y.; Liao, Q.~V.; Sattigeri, P.; Fogliato,
  R.; Melan{\c{c}}on, G.; Krishnan, R.; Stanley, J.; Tickoo, O.; et~al. 2021.
\newblock Uncertainty as a form of transparency: Measuring, communicating, and
  using uncertainty.
\newblock In \emph{Proceedings of the 2021 AAAI/ACM Conference on AI, Ethics,
  and Society}, 401--413.

\bibitem[{Borisov et~al.(2021)Borisov, Leemann, Se{\ss}ler, Haug, Pawelczyk,
  and Kasneci}]{survey_DL_tabular}
Borisov, V.; Leemann, T.; Se{\ss}ler, K.; Haug, J.; Pawelczyk, M.; and Kasneci,
  G. 2021.
\newblock Deep Neural Networks and Tabular Data: {A} Survey.
\newblock \emph{CoRR}, abs/2110.01889.

\bibitem[{Carlini and Wagner(2017)}]{adversarialUncertainty}
Carlini, N.; and Wagner, D. 2017.
\newblock Adversarial examples are not easily detected: Bypassing ten detection
  methods.
\newblock In \emph{Proceedings of the 10th ACM workshop on artificial
  intelligence and security}, 3--14.

\bibitem[{Chen et~al.(2020)Chen, Janizek, Lundberg, and
  Lee}]{true_to_the_model_true_to_the_data}
Chen, H.; Janizek, J.~D.; Lundberg, S.~M.; and Lee, S. 2020.
\newblock True to the Model or True to the Data?
\newblock \emph{CoRR}, abs/2006.16234.

\bibitem[{Chen and Guestrin(2016)}]{xgboost}
Chen, T.; and Guestrin, C. 2016.
\newblock Xgboost: A scalable tree boosting system.
\newblock In \emph{Proceedings of the 22nd acm sigkdd international conference
  on knowledge discovery and data mining}, 785--794.

\bibitem[{Cortes et~al.(2008)Cortes, Mohri, Riley, and
  Rostamizadeh}]{cortes2008sample}
Cortes, C.; Mohri, M.; Riley, M.; and Rostamizadeh, A. 2008.
\newblock Sample selection bias correction theory.
\newblock In \emph{International conference on algorithmic learning theory},
  38--53. Springer.

\bibitem[{D'Angelo and Henning(2021)}]{dangelo2021uncertaintybased}
D'Angelo, F.; and Henning, C. 2021.
\newblock Uncertainty-based out-of-distribution detection requires suitable
  function space priors.
\newblock \emph{arXiv preprint arXiv:2110.06020}.

\bibitem[{Diethe et~al.(2019)Diethe, Borchert, Thereska, Balle, and
  Lawrence}]{continual_learning}
Diethe, T.; Borchert, T.; Thereska, E.; Balle, B.; and Lawrence, N. 2019.
\newblock Continual Learning in Practice.
\newblock \emph{stat}, 1050: 18.

\bibitem[{Dua and Graff(2017)}]{uci_data}
Dua, D.; and Graff, C. 2017.
\newblock {UCI} Machine Learning Repository.
\newblock University of California, Irvine, School of Information and Computer
  Sciences, \url{http://archive.ics.uci.edu/ml}.

\bibitem[{Elsayed et~al.(2021)Elsayed, Thyssens, Rashed, Schmidt{-}Thieme, and
  Jomaa}]{do_we_need_DL}
Elsayed, S.; Thyssens, D.; Rashed, A.; Schmidt{-}Thieme, L.; and Jomaa, H.~S.
  2021.
\newblock Do We Really Need Deep Learning Models for Time Series Forecasting?
\newblock \emph{CoRR}, abs/2101.02118.

\bibitem[{Friedman et~al.(2001)Friedman, Hastie, Tibshirani
  et~al.}]{statisticallearning}
Friedman, J.; Hastie, T.; Tibshirani, R.; et~al. 2001.
\newblock \emph{The elements of statistical learning}, volume 1.10.
\newblock Springer series in statistics New York.

\bibitem[{Gal and Ghahramani(2016)}]{gal2016dropout}
Gal, Y.; and Ghahramani, Z. 2016.
\newblock Dropout as a bayesian approximation: Representing model uncertainty
  in deep learning.
\newblock In \emph{international conference on machine learning}, 1050--1059.
  PMLR.

\bibitem[{Garg et~al.(2021)Garg, Balakrishnan, Lipton, Neyshabur, and
  Sedghi}]{garg2022leveraging}
Garg, S.; Balakrishnan, S.; Lipton, Z.~C.; Neyshabur, B.; and Sedghi, H. 2021.
\newblock Leveraging Unlabeled Data to Predict Out-of-Distribution Performance.
\newblock In \emph{NeurIPS 2021 Workshop on Distribution Shifts: Connecting
  Methods and Applications}.

\bibitem[{Grinsztajn, Oyallon, and Varoquaux(2022)}]{grinsztajn:hal-03723551}
Grinsztajn, L.; Oyallon, E.; and Varoquaux, G. 2022.
\newblock {Why do tree-based models still outperform deep learning on tabular
  data?}
\newblock Working paper or preprint.

\bibitem[{Guschin et~al.(2018)Guschin, Ulyanov, Trofimov, Altukhov, and
  Michaidilis}]{howtowinKaggle}
Guschin, A.; Ulyanov, D.; Trofimov, M.; Altukhov, D.; and Michaidilis, M. 2018.
\newblock How to Win a Data Science Competition: Learn from Top Kagglers -
  National Research University Higher School of Economics.
\newblock
  \url{https://www.coursera.org/lecture/competitive-data-science/categorical-and-ordinal-features-qu1TF}.
\newblock Accessed 02/11/20.

\bibitem[{He et~al.(2014)He, Pan, Jin, Xu, Liu, Xu, Shi, Atallah, Herbrich,
  Bowers et~al.}]{practicalFB}
He, X.; Pan, J.; Jin, O.; Xu, T.; Liu, B.; Xu, T.; Shi, Y.; Atallah, A.;
  Herbrich, R.; Bowers, S.; et~al. 2014.
\newblock Practical lessons from predicting clicks on ads at facebook.
\newblock In \emph{Proceedings of the Eighth International Workshop on Data
  Mining for Online Advertising}, 1--9.

\bibitem[{Heckman(1990)}]{varietiesSelectionBias}
Heckman, J. 1990.
\newblock Varieties of selection bias.
\newblock \emph{The American Economic Review}, 80(2): 313--318.

\bibitem[{Huang et~al.(2006)Huang, Gretton, Borgwardt, Sch{\"o}lkopf, and
  Smola}]{CorrectingSampleSelection}
Huang, J.; Gretton, A.; Borgwardt, K.; Sch{\"o}lkopf, B.; and Smola, A. 2006.
\newblock Correcting sample selection bias by unlabeled data.
\newblock \emph{Advances in neural information processing systems}, 19:
  601--608.

\bibitem[{Jiang et~al.(2021)Jiang, Nagarajan, Baek, and
  Kolter}]{jiang2021assessing}
Jiang, Y.; Nagarajan, V.; Baek, C.; and Kolter, J.~Z. 2021.
\newblock Assessing generalization of sgd via disagreement.
\newblock \emph{arXiv preprint arXiv:2106.13799}.

\bibitem[{Kirsch, Van~Amersfoort, and Gal(2019)}]{kirsch2019batchbald}
Kirsch, A.; Van~Amersfoort, J.; and Gal, Y. 2019.
\newblock Batchbald: Efficient and diverse batch acquisition for deep bayesian
  active learning.
\newblock \emph{Advances in neural information processing systems}, 32:
  7026--7037.

\bibitem[{Koh and Liang(2017)}]{koh2020understanding}
Koh, P.~W.; and Liang, P. 2017.
\newblock Understanding black-box predictions via influence functions.
\newblock In \emph{International Conference on Machine Learning}, 1885--1894.
  PMLR.

\bibitem[{Kumar and Srivastava(2012)}]{kumar2012bootstrap}
Kumar, S.; and Srivastava, A. 2012.
\newblock Bootstrap prediction intervals in non-parametric regression with
  applications to anomaly detection.
\newblock In \emph{Proc. 18th ACM SIGKDD Conf. Knowl. Discovery Data Mining}.

\bibitem[{Lakshminarayanan(2021)}]{unc_ood_class}
Lakshminarayanan, B. 2021.
\newblock Uncertainty and Out-of-Distribution Robustness in Deep Learning.

\bibitem[{Lakshminarayanan, Pritzel, and
  Blundell(2017)}]{lakshminarayanan2017simple}
Lakshminarayanan, B.; Pritzel, A.; and Blundell, C. 2017.
\newblock Simple and Scalable Predictive Uncertainty Estimation using Deep
  Ensembles.
\newblock \emph{Advances in Neural Information Processing Systems}, 30.

\bibitem[{Lundberg et~al.(2020{\natexlab{a}})Lundberg, Erion, Chen, DeGrave,
  Prutkin, Nair, Katz, Himmelfarb, Bansal, and Lee}]{lundberg2020local2global}
Lundberg, S.~M.; Erion, G.; Chen, H.; DeGrave, A.; Prutkin, J.~M.; Nair, B.;
  Katz, R.; Himmelfarb, J.; Bansal, N.; and Lee, S.-I. 2020{\natexlab{a}}.
\newblock From local explanations to global understanding with explainable AI
  for trees.
\newblock \emph{Nature machine intelligence}, 2(1): 56--67.

\bibitem[{Lundberg et~al.(2020{\natexlab{b}})Lundberg, Erion, Chen, DeGrave,
  Prutkin, Nair, Katz, Himmelfarb, Bansal, and Lee}]{shapTree}
Lundberg, S.~M.; Erion, G.~G.; Chen, H.; DeGrave, A.~J.; Prutkin, J.~M.; Nair,
  B.; Katz, R.; Himmelfarb, J.; Bansal, N.; and Lee, S. 2020{\natexlab{b}}.
\newblock From local explanations to global understanding with explainable {AI}
  for trees.
\newblock \emph{Nat. Mach. Intell.}, 2(1): 56--67.

\bibitem[{Lundberg and Lee(2017)}]{lundberg2017unified}
Lundberg, S.~M.; and Lee, S.-I. 2017.
\newblock A unified approach to interpreting model predictions.
\newblock In \emph{Proceedings of the 31st international conference on neural
  information processing systems}, 4768--4777.

\bibitem[{Lundberg et~al.(2018)Lundberg, Nair, Vavilala, Horibe, Eisses, Adams,
  Liston, Low, Newman, Kim et~al.}]{lundberg2018}
Lundberg, S.~M.; Nair, B.; Vavilala, M.~S.; Horibe, M.; Eisses, M.~J.; Adams,
  T.; Liston, D.~E.; Low, D. K.-W.; Newman, S.-F.; Kim, J.; et~al. 2018.
\newblock Explainable machine-learning predictions for the prevention of
  hypoxaemia during surgery.
\newblock \emph{Nature biomedical engineering}, 2(10): 749--760.

\bibitem[{Malinin(2019)}]{malinin2019uncertainty}
Malinin, A. 2019.
\newblock \emph{Uncertainty estimation in deep learning with application to
  spoken language assessment}.
\newblock Ph.D. thesis, University of Cambridge.

\bibitem[{Malinin et~al.(2021)Malinin, Band, Gal, Gales, Ganshin, Chesnokov,
  Noskov, Ploskonosov, Prokhorenkova, Provilkov et~al.}]{ShiftsData}
Malinin, A.; Band, N.; Gal, Y.; Gales, M.; Ganshin, A.; Chesnokov, G.; Noskov,
  A.; Ploskonosov, A.; Prokhorenkova, L.; Provilkov, I.; et~al. 2021.
\newblock Shifts: A Dataset of Real Distributional Shift Across Multiple
  Large-Scale Tasks.
\newblock In \emph{Thirty-fifth Conference on Neural Information Processing
  Systems Datasets and Benchmarks Track (Round 2)}.

\bibitem[{Mougan, Kanellos, and Gottron(2021)}]{desiderataECB}
Mougan, C.; Kanellos, G.; and Gottron, T. 2021.
\newblock Desiderata for Explainable {AI} in Statistical Production Systems of
  the European Central Bank.
\newblock In \emph{Machine Learning and Principles and Practice of Knowledge
  Discovery in Databases - International Workshops of {ECML} {PKDD} 2021,
  Virtual Event, September 13-17, 2021, Proceedings, Part {I}}, volume 1524 of
  \emph{Communications in Computer and Information Science}, 575--590.
  Springer.

\bibitem[{Neyshabur et~al.(2017)Neyshabur, Bhojanapalli, Mcallester, and
  Srebro}]{neyshabur2017exploring}
Neyshabur, B.; Bhojanapalli, S.; Mcallester, D.; and Srebro, N. 2017.
\newblock Exploring Generalization in Deep Learning.
\newblock \emph{Advances in Neural Information Processing Systems}, 30:
  5947--5956.

\bibitem[{Neyshabur et~al.(2019)Neyshabur, Li, Bhojanapalli, LeCun, and
  Srebro}]{neyshabur2018understanding}
Neyshabur, B.; Li, Z.; Bhojanapalli, S.; LeCun, Y.; and Srebro, N. 2019.
\newblock Towards Understanding the Role of Over-Parametrization in
  Generalization of Neural Networks.
\newblock In \emph{International Conference on Learning Representations
  (ICLR)}.

\bibitem[{Ovadia et~al.(2019)Ovadia, Fertig, Ren, Nado, Sculley, Nowozin,
  Dillon, Lakshminarayanan, and Snoek}]{trustUncertainty}
Ovadia, Y.; Fertig, E.; Ren, J.; Nado, Z.; Sculley, D.; Nowozin, S.; Dillon,
  J.~V.; Lakshminarayanan, B.; and Snoek, J. 2019.
\newblock Can You Trust Your Model’s Uncertainty? Evaluating Predictive
  Uncertainty Under Dataset Shift.
\newblock \emph{stat}, 1050: 17.

\bibitem[{Pedregosa et~al.(2011)Pedregosa, Varoquaux, Gramfort, Michel,
  Thirion, Grisel, Blondel, Prettenhofer, Weiss, Dubourg
  et~al.}]{pedregosa2011scikit}
Pedregosa, F.; Varoquaux, G.; Gramfort, A.; Michel, V.; Thirion, B.; Grisel,
  O.; Blondel, M.; Prettenhofer, P.; Weiss, R.; Dubourg, V.; et~al. 2011.
\newblock Scikit-learn: Machine learning in Python.
\newblock \emph{the Journal of machine Learning research}, 12: 2825--2830.

\bibitem[{Qui{\~n}onero-Candela et~al.(2008)Qui{\~n}onero-Candela, Sugiyama,
  Schwaighofer, and Lawrence}]{datasetShift}
Qui{\~n}onero-Candela, J.; Sugiyama, M.; Schwaighofer, A.; and Lawrence, N.~D.
  2008.
\newblock \emph{Dataset shift in machine learning}.
\newblock Mit Press.

\bibitem[{Rabanser, G{\"{u}}nnemann, and
  Lipton(2019)}]{DBLP:conf/nips/RabanserGL19}
Rabanser, S.; G{\"{u}}nnemann, S.; and Lipton, Z.~C. 2019.
\newblock Failing Loudly: An Empirical Study of Methods for Detecting Dataset
  Shift.
\newblock In Wallach, H.~M.; Larochelle, H.; Beygelzimer, A.;
  d'Alch{\'{e}}{-}Buc, F.; Fox, E.~B.; and Garnett, R., eds., \emph{Advances in
  Neural Information Processing Systems 32: Annual Conference on Neural
  Information Processing Systems 2019, NeurIPS 2019, December 8-14, 2019,
  Vancouver, BC, Canada}, 1394--1406.

\bibitem[{Ribeiro, Singh, and Guestrin(2016)}]{ribeiro2016why}
Ribeiro, M.~T.; Singh, S.; and Guestrin, C. 2016.
\newblock " Why should i trust you?" Explaining the predictions of any
  classifier.
\newblock In \emph{Proceedings of the 22nd ACM SIGKDD international conference
  on knowledge discovery and data mining}, 1135--1144.

\bibitem[{Shimodaira(2000)}]{SHIMODAIRA2000227}
Shimodaira, H. 2000.
\newblock Improving predictive inference under covariate shift by weighting the
  log-likelihood function.
\newblock \emph{Journal of statistical planning and inference}, 90(2):
  227--244.

\bibitem[{Smith and Gal(2018)}]{smith2018understanding}
Smith, L.; and Gal, Y. 2018.
\newblock Understanding Measures of Uncertainty for Adversarial Example
  Detection.
\newblock In Globerson, A.; and Silva, R., eds., \emph{Proceedings of the
  Thirty-Fourth Conference on Uncertainty in Artificial Intelligence, {UAI}
  2018, Monterey, California, USA, August 6-10, 2018}, 560--569. {AUAI} Press.

\bibitem[{Stolzenberg and Relles(1997)}]{intuitionSampleSelection}
Stolzenberg, R.~M.; and Relles, D.~A. 1997.
\newblock Tools for intuition about sample selection bias and its correction.
\newblock \emph{American sociological review}, 494--507.

\bibitem[{Sugiyama, Krauledat, and M{\"u}ller(2007)}]{sugiyama}
Sugiyama, M.; Krauledat, M.; and M{\"u}ller, K.-R. 2007.
\newblock Covariate shift adaptation by importance weighted cross validation.
\newblock \emph{Journal of Machine Learning Research}, 8(5).

\bibitem[{Sugiyama and M{\"u}ller(2005)}]{sugiyama2}
Sugiyama, M.; and M{\"u}ller, K.-R. 2005.
\newblock Input-dependent estimation of generalization error under covariate
  shift.
\newblock \emph{.}

\bibitem[{Sundararajan, Taly, and Yan(2017)}]{sundararajan2017axiomatic}
Sundararajan, M.; Taly, A.; and Yan, Q. 2017.
\newblock Axiomatic attribution for deep networks.
\newblock In \emph{International Conference on Machine Learning}, 3319--3328.
  PMLR.

\bibitem[{Tasche(2017)}]{priorShift}
Tasche, D. 2017.
\newblock Fisher consistency for prior probability shift.
\newblock \emph{The Journal of Machine Learning Research}, 18(1): 3338--3369.

\bibitem[{Zadrozny(2004)}]{learningSampleSelection}
Zadrozny, B. 2004.
\newblock Learning and evaluating classifiers under sample selection bias.
\newblock In \emph{Proceedings of the twenty-first international conference on
  Machine learning}, 114.

\end{thebibliography}

\pagebreak
%\section*{Appendix}
\section{Synthetic data experiments}
In this section we apply our model monitoring method to synthetic data to demonstrate main differences between our approach, methods from classical statistics and other model agnostic uncertainty approaches.

We sample data from a three-variate normal distribution $X = (X_1,X_2,X_3) \sim N(1,0.1\cdot I_3)$ with $I_3$ being an identity matrix of order 3. We construct the target variable as $Y:=X_1^2 + X_2 + \varepsilon$, with $\varepsilon \sim N(0,0.1)$ being random noise. We thus have a non-linear feature $X_1^2$, a linear feature $X_2$ and a feature $X_3$ which is not used, all of them being independent among each other. We draw $10,000$ random samples for both training and test data, yielding the training set $(X^{tr},Y^{tr})$ and the test set $(X^{te},Y^{te})$ and train a linear model $f_\theta$.

\subsection{Evaluating model deterioration}
In this experiment we aim to compare indicators of model deterioration by replacing the value of each feature, $j=1,2,3$, with a continuous vector with evenly spaced numbers over a specified range $(-3,4)$

\begin{table}[h]
\begin{tabular}{ll}
\hline
Notation                    & Meaning                                                           \\\hline
$y$                         & True label/model.                                                 \\
$d$                         & Dimension of the feature space.                                   \\
$\delta$                    & Deterministic component of $y$.                                   \\
$\varepsilon$               & Observation noise component of $y$.                               \\
$X$                         & Our data sample.                                                  \\
$N$                         & Size of $X$.                                                       \\
$\hat{\delta}^{(N)}$        & Finite sample estimate of $\delta$.                               \\
$\hat{\delta}$              & $\lim_{n\to\infty}\hat{\delta}^{(n)}$                             \\
$\beta$                     & The bias function.                                                \\
$\varepsilon_v^{(N)}$       & Model variance noise.                                             \\
$B$                         & Number of bootstrap samples.                                      \\
$b$                         & The index of a particular bootstrap sample.                       \\
$\overline{\delta}_b^{(N)}$ & Bootstrap estimate of $\delta$.                                   \\
$m_b^{(N)}$                 & Centered bootstrap estimate of $\delta$.                          \\
$\mathbb{E}[Y]$             & Expected value of $Y$.                                            \\
$\mathbb{D}[Y]$            & Distribution function (CDF) of $Y$.                               \\
$q_\xi(Y)$                  & The $\xi$'th quantile of $Y$.                                     \\\hline
\end{tabular}
\caption{Notation used throughout the paper.}\label{tab:notation}
\end{table}

\begin{figure}[ht]
  \includegraphics[width=1\linewidth]{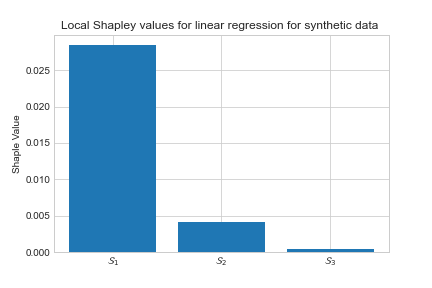}
  \caption{Shapley values for the data point $x:=(10,10,10)$  and the model $g_\theta$. Explainable uncertainty estimation allows to
  account for the drivers of uncertainty that serves as a proxy for model predictive performance degradation. }\label{fig:syntheticAnalytical}
\end{figure}

\begin{figure*}[ht]
  \includegraphics[width=1\linewidth]{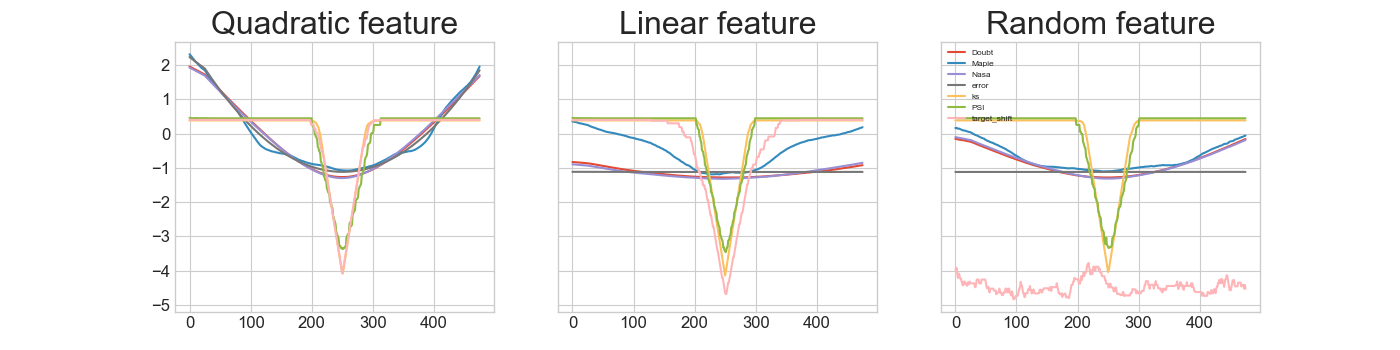}
  \caption{Comparison of different monitoring methods over the synthetic dataset. Each of the plots represents an independent experiment where each of the features has been replaced by a continuous vector with  evenly spaced numbers  in the range $(-3,4)$.  The x-axis represents the index of the sorted column. Doubt achieves a better goodness-of-fit than previous statistical methods (cf. Table \ref{table:synthetic}) }\label{fig:syntheticDegradation}
\end{figure*}

\begin{table*}[ht]
\begin{center}
\begin{tabular}{c|cccccc}
                  & \textbf{Doubt} & \textbf{NASA}  & \textbf{MAPIE}   & \textbf{KS} & \textbf{PSI} & \textbf{Prediction Shift} \\ \hline
Quadratic feature &  \textbf{0.09}&     0.11        &    \textbf{0.09}&     0.96    &0.95          &   0.94                  \\
Linear feature    &  \textbf{0.11}&     0.12        &     0.70       &     0.35     & 1.39         &  1.46                     \\
Random feature    &  \textbf{0.34}&     0.35        &     0.42       &     1.39     &1.46          &   5.58                  \\ \hline
Mean              & \textbf{0.18} &     0.20        &     0.40       &      1.24    &1.29          &   2.69
\end{tabular}
\caption{Performance of model monitoring systems for model deterioration for a linear regression model on a synthetic dataset (cf. Figure \ref{fig:syntheticDegradation}).  K-S and PSI are the monitoring systems obtained by computing the Kolmogorov-Smirnov test values and the Population Stability Index, respectively, and Doubt is our method. Best results are shown in bold.}\label{table:synthetic}
\end{center}
\end{table*}

In Figure \ref{fig:syntheticDegradation} and Table \ref{table:synthetic} we can see what is the impact of the synthetic distribution shift is in terms of model mean squared error. We see that our uncertainty method follows a better goodness-of-fit than both the Kolmogorov-Smirnov test and the Population Stability Index on the input data and and target data. The latter two methods have identical values for the linear and random features since they are independent of the target values.

\subsection{Detecting the source of uncertainty/model deterioration}

For this example, we simulate out-of-distribution data by shifting $X$ by 10; i.e., $X^{ood}_j := X^{tr}_j + 10$ for $j=1,2,3$. We train a linear regression model $f_\theta$ on $(X^{tr},Y^{tr})$ and use Doubt to get uncertainty estimates $Z$ on both $X^{te}$ and $X^{ood}$. We then train another linear model $g_\psi$ on $((X^{te},X^{ood}))$ to predict $Z$. The coefficients of $g_\psi$ are $\beta_1= 0.03478, \beta_2=0.005023 ,\beta_3= 0.000522$.
Since the features are independent and we are dealing with linear regression, the interventional conditional expectation Shapley values can be computed as $\beta_i(x_i-\mu_i)$ \cite{true_to_the_model_true_to_the_data}, where $\mu$ is the mean of the training data . So for the data point $x:=(10,10,10)$, the Shapley values are $(0.0291,0.0042,0.0004)$, where the most relevant shifted feature in the model is the one that receives the highest Shapley value. In this experiment with synthetic data, statistical testing on the input data would have flagged the three feature distributions as equally shifted. With our proposed method, we can identify the more meaningful features towards identifying the source of model predictive performance deterioration. It is worth noting that this explainable AI uncertainty approach can be used with other uncertainty estimation techniques.

\section{Computational performance comparison}

\begin{figure*}[ht]
\centering
\includegraphics[width=.3\textwidth]{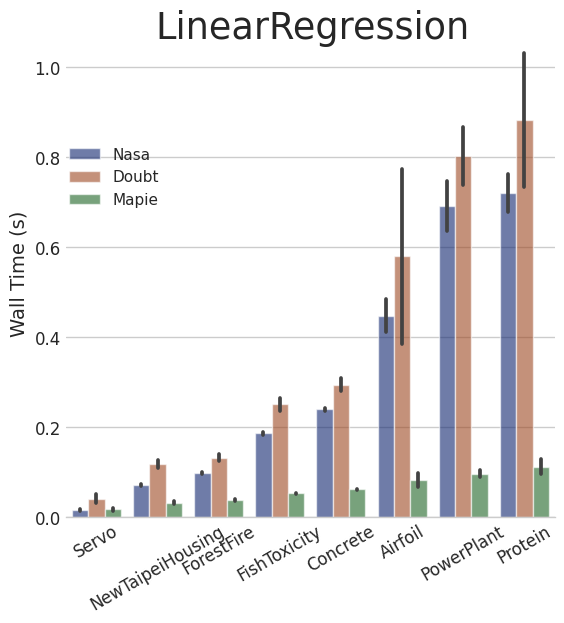}\hfill
\includegraphics[width=.3\textwidth]{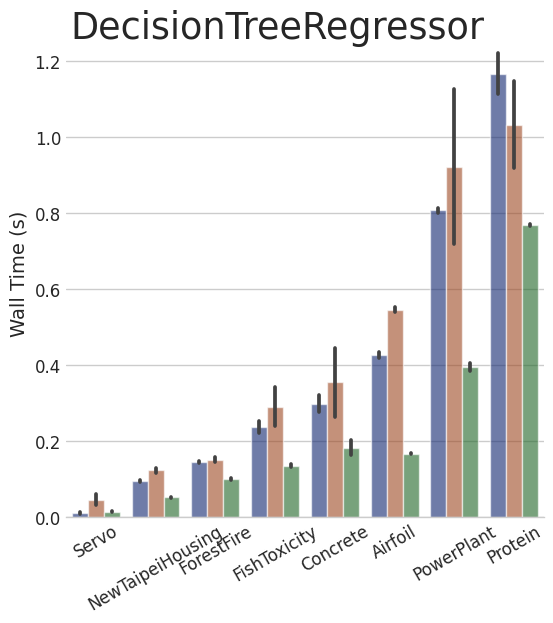}\hfill
\includegraphics[width=.3\textwidth]{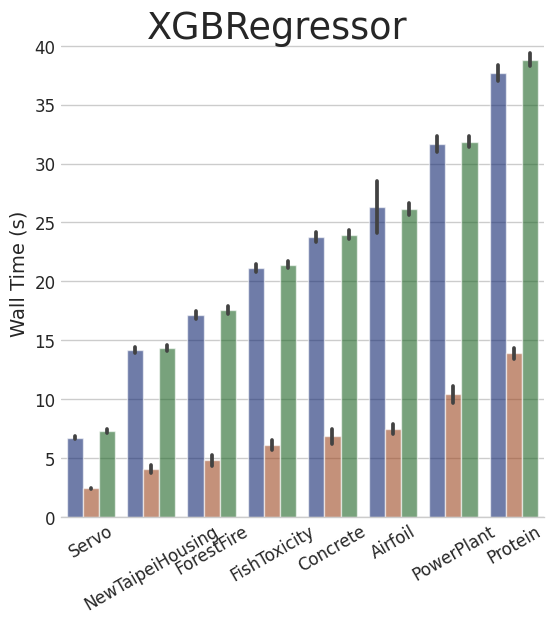}

\caption{Wall time computational performance of the three uncertainty estimation methods described through the paper. Datasets are sorted by the number of rows and the y-axis is not shared accross plots.}
\label{fig:figure3}
\end{figure*}

In this section we compare the wall clock time of the NASA method \cite{kumar2012bootstrap}, MAPIE \cite{barber2021predictive} and Doubt with three estimators: linear regression, decision tree and  gradient boosting decision tree, over the range of dataset previously used. We set the number of model bootstraps in the Doubt method to be equal to the number of cross-validation splits in the MAPIE method; namely, the square root of the total number of samples in the dataset. The experiment are performed on a virtual machine with 8 vCPUs and 60 GB RAM.

See the results in Figure \ref{fig:figure3}. We see here that the MAPIE approach is significantly faster than the NASA and Doubt methods in the cases where the model used is a linear regression model or a decision tree. However, interestingly, in the case where the model is an XGBoost model, the Doubt method is significantly faster than the other two.

\begin{figure*}[ht]
  \centering
  \includegraphics[width=1\linewidth]{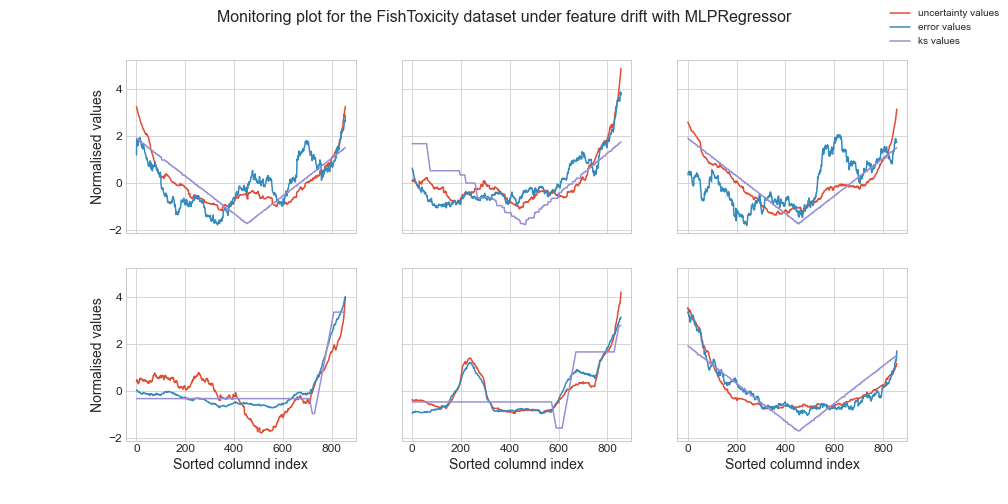}
  \caption{Enlarged version of Figure \ref{fig:distribution}. Comparison of different model degradation detection methods for the Fish Toxicity dataset. Each of the plots represents an independent experiment where each of the six features has a sorted shift. Doubt achieves a better goodness-of-fit than previous statistical methods.}
  \label{fig:distributionBIG}
\end{figure*}

\begin{landscape}
\begin{table}[ht]
\begin{tabular}{l|l|llllllll|ll}
                            &             & Airfoil & Concrete & ForestFire & Parkinsons & PowerPlant & Protein & Bike & FishToxicity & Mean               & Std                 \\\hline
Linear Regression           & Uncertainty & 0.67    & 0.59     & 0.6        & 0.6        & 1.0        & 0.72    & 0.86 & 0.64         & 0.71               & 0.14  \\
                            & MAPIE       & 0.72    & 0.79     & 0.71       & 0.62       & 0.85       & 0.74    & 0.79 & 0.94         & 0.77               & 0.09 \\
                            & K-S         & 0.78    & 0.67     & 0.81       & 0.92       & 0.8        & 0.99    & 0.85 & 0.71         & 0.81               & 0.10 \\
                            & PSI         & 1.02    & 0.9      & 0.78       & 0.83       & 0.93       & 0.79    & 0.91 & 0.81         & 0.87               & 0.08 \\
                            & Target      & 0.94    & 0.68     & 0.7        & 0.84       & 0.8        & 0.97    & 0.97 & 1.0          & 0.86               & 0.12 \\\hline
Poisson Regressor           & Uncertainty & 0.81    & 0.49     & 0.93       & 0.82       & 0.97       & 0.83    & 0.79 & 0.71         & 0.79               & 0.14 \\
                            & MAPIE       & 0.84    & 0.55     & 0.89       & 0.86       & 0.87       & 0.91    & 0.61 & 1.1          & 0.82               & 0.17 \\
                            & K-S         & 0.83    & 0.68     & 1.08       & 1.08       & 0.8        & 1.16    & 1.16 & 0.79         & 0.94               & 0.19 \\
                            & PSI         & 1.01    & 0.9      & 1.04       & 1.04       & 0.94       & 0.85    & 0.85 & 0.86         & 0.93               & 0.08 \\
                            & Target      & 1.02    & 0.75     & 1.15       & 1.15       & 0.79       & 1.1     & 1.1  & 0.95         & 1.00               & 0.15  \\\hline
Decision Tree (20)          & Uncertainty & 0.48    & 0.44     & 0.74       & 0.51       & 0.45       & 0.44    & 0.46 & 0.47         & 0.49               & 0.10 \\
                            & MAPIE       & 0.59    & 0.52     & 0.96       & 0.61       & 0.61       & 0.51    & 0.49 & 0.45         & 0.59               & 0.15 \\
                            & K-S         & 0.49    & 0.56     & 0.66       & 0.49       & 0.3        & 0.49    & 0.68 & 0.5          & 0.52               & 0.11  \\
                            & PSI         & 1.14    & 1.08     & 1.02       & 0.9        & 0.92       & 0.9     & 0.86 & 0.95         & 0.97               & 0.09  \\
                            & Target      & 0.96    & 0.77     & 0.81       & 0.9        & 0.64       & 0.54    & 0.94 & 0.79         & 0.79               & 0.14 \\\hline
Random Forest Regressor     & Uncertainty & 0.57    & 0.577    & 0.93       & 0.5        & 0.8        & 0.92    & 0.67 & 0.98         & 0.74               & 0.18 \\
                            & MAPIE       & 0.76    & 0.91     & 0.95       & 0.58       & 0.99       & 1.0     & 0.75 & 0.98         & 0.86               & 0.15 \\
                            & K-S         & 0.48    & 0.51     & 0.74       & 0.41       & 0.41       & 0.41    & 0.62 & 0.45         & 0.50               & 0.11 \\
                            & PSI         & 1.11    & 1.02     & 1.0        & 0.89       & 0.92       & 0.92    & 0.85 & 0.93         & 0.95               & 0.08 \\
                            & Target      & 0.95    & 0.7      & 0.7        & 0.87       & 0.61       & 0.38    & 0.89 & 0.8          & 0.73               & 0.18 \\\hline
Gradient Boosting Regressor & Uncertainty & 0.48    & 0.49     & 1.0        & 0.45       & 0.45       & 0.32    & 0.57 & 0.88         & 0.58               & 0.23 \\
                            & MAPIE       & 0.66    & 0.86     & 0.94       & 0.55       & 0.67       & 0.51    & 0.64 & 1.03         & 0.73               & 0.18 \\
                            & K-S         & 0.45    & 0.47     & 0.65       & 0.9        & 0.9        & 0.42    & 0.64 & 0.48         & 0.61               & 0.19 \\
                            & PSI         & 1.11    & 1.01     & 0.99       & 0.91       & 0.92       & 0.9     & 0.85 & 0.98         & 0.95               & 0.08 \\
                            & Target      & 1.05    & 0.73     & 0.54       & 0.92       & 0.65       & 0.46    & 0.88 & 0.8          & 0.75               & 0.19 \\\hline
MultiLayer-Perceptron       & Uncertainty & 1.5     & 0.93     & 0.81       & 0.43       & 0.43       & 0.42    & 0.52 & 0.41         & 0.68               & 0.38 \\
                            & MAPIE       & 1.6     & 0.85     & 0.95       & 0.49       & 0.58       & 0.42    & 0.55 & 0.55         & 0.74               & 0.38  \\
                            & K-S         & 1.03    & 0.72     & 1.18       & 0.61       & 0.62       & 0.64    & 0.62 & 0.62         & 0.75               & 0.22 \\
                            & PSI         & 1.1     & 0.75     & 1.08       & 0.76       & 0.88       & 0.79    & 0.74 & 0.66         & 0.84               & 0.16  \\
                            & Target      & 0.63    & 0.8      & 1.24       & 0.6        & 0.59       & 0.54    & 0.73 & 0.78         & 0.73               & 0.22
\end{tabular}
\caption{Non-aggregated version of Table \ref{tab:scores}. Model monitoring systems for model deterioration for a variety of model architectures on eight regression datasets from the UCI repository \cite{uci_data}.}
\label{tab:appendix}
\end{table}
\end{landscape}

\end{document}